\documentclass[final]{cvpr}
\usepackage{times}
\usepackage{epsfig}
\usepackage{graphicx}
\usepackage{amsmath}
\usepackage{amssymb}
\usepackage{xcolor}
\usepackage{paralist}
\usepackage{tabularx}
\usepackage{booktabs}
\usepackage{array,etoolbox}
\preto\tabular{\setcounter{magicrownumbers}{0}}
\newcounter{magicrownumbers}

\usepackage[T1]{fontenc}
\usepackage[utf8]{inputenc}

\usepackage[pagebackref=true,breaklinks=true,letterpaper=true,colorlinks,bookmarks=false]{hyperref}
\def\confYear{CVPR 2021}
\providecommand{\ModelName}{KRISP\xspace}
\providecommand{\ModelNameLong}{Knowledge Reasoning with Implicit and  Symbolic rePresentations\xspace}
\providecommand{\MMBERTBase}{Multi-modal BERT\xspace}
\providecommand{\MMBERTLong}{Multi-modal BERT\xspace}
\providecommand{\MMBERTAbr}{MMBERT\xspace}
\providecommand{\myparagraph}[1]{\noindent\textbf{#1.}}

\providecommand{\sectionvspace}{\vspace{-0cm}}

\begin{document}

\title{KRISP: Integrating Implicit and Symbolic Knowledge for Open-Domain Knowledge-Based VQA}

\author{Kenneth Marino$^{1,2}$ \hspace{-.4cm} \\
\and
Xinlei Chen$^2$ \hspace{-.4cm} \\
\and
Devi Parikh$^{2,3}$ \hspace{-.4cm}\\
\and
Abhinav Gupta$^{1,2}$ \hspace{-.4cm}\\
\and
Marcus Rohrbach$^1$\\
\and
$^1$Carnegie Mellon University
\and
$^2$Facebook AI Research
\and
$^3$Georgia Tech
}
\maketitle

\begin{abstract}
    One of the most challenging question types in VQA is when  answering the question requires outside knowledge not present in the image. In this work we study open-domain knowledge, the setting when the knowledge required to answer a question is not given/annotated, neither at training nor test time. We tap into two types of knowledge representations and reasoning. First, implicit knowledge which can be learned effectively from unsupervised language pre-training and supervised training data with transformer-based models. Second, explicit, symbolic knowledge encoded in knowledge bases. Our approach combines both---exploiting the powerful implicit reasoning of transformer models for answer prediction, and integrating symbolic representations from a knowledge graph, while never losing their explicit semantics to an implicit embedding. We combine diverse sources of knowledge to cover the wide variety of knowledge needed to solve knowledge-based questions. We show our approach, \emph{\ModelName (\ModelNameLong)}, significantly outperforms state-of-the-art on OK-VQA, the largest available dataset for open-domain knowledge-based VQA. We show with extensive ablations that while our model successfully exploits implicit knowledge reasoning, the symbolic answer module which explicitly connects the knowledge graph to the answer vocabulary is critical to the performance of our method and generalizes to rare answers.
\end{abstract}

\vspace{-.3cm}
\sectionvspace
\section{Introduction}
\sectionvspace

\begin{figure}[t]
\centering
\includegraphics[width=0.9\linewidth]{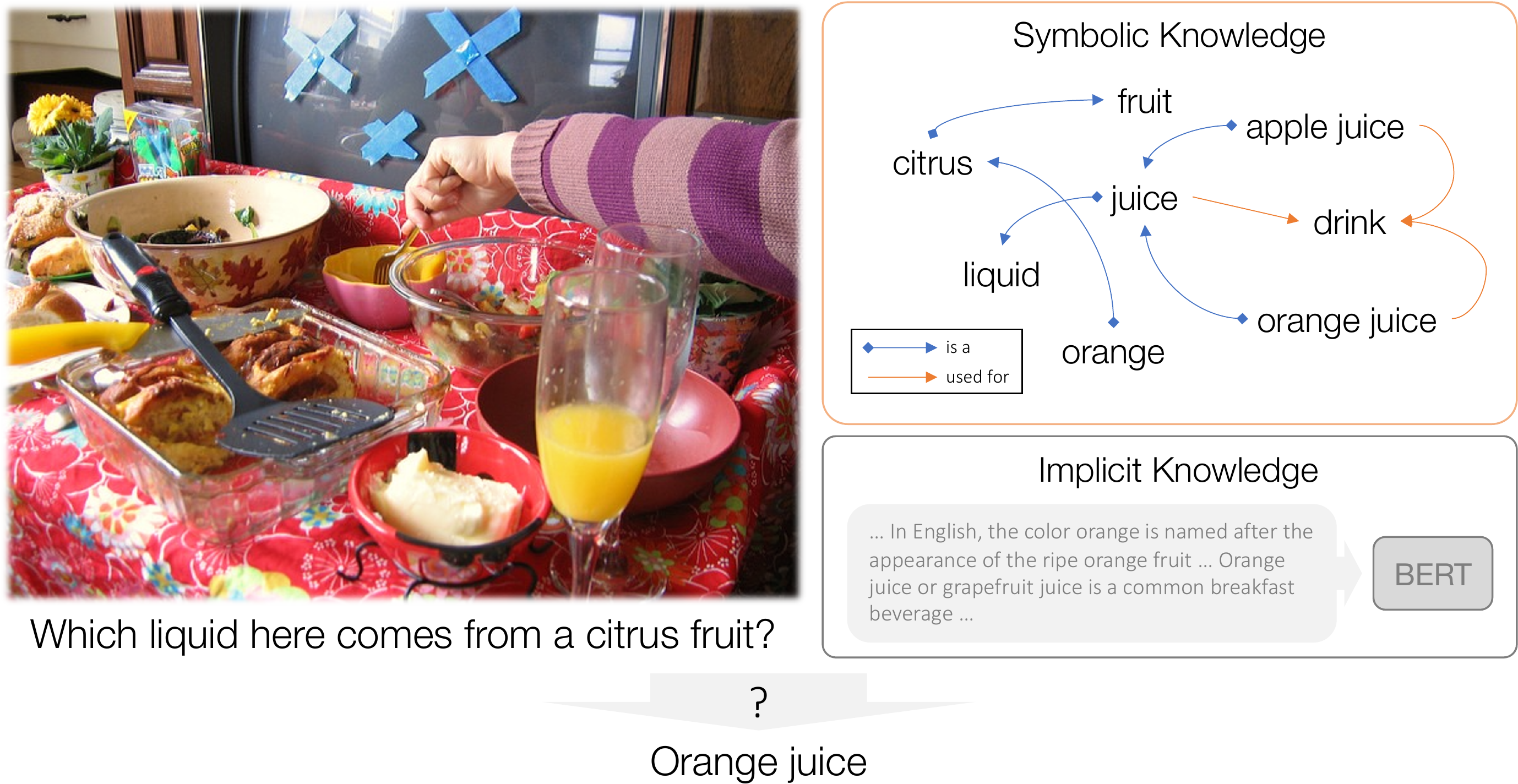}
\caption{An OK-VQA example that requires external knowledge.
Our \ModelName model 
uses a symbolic knowledge graph as well as the implicit knowledge learned from large-scale BERT training to answer the question.}
\label{fig:teaser}
\end{figure}

Consider the example shown from a recent VQA dataset~\cite{marino19cvpr} in Fig.~\ref{fig:teaser}. 
To answer this question, we not only need to parse the question and understand the image but also use external knowledge. Early work in VQA focused on image and question parsing~
\cite{agrawal17, antol15, fukui16, malinowski14a, malinowski15} assuming  all required knowledge can be learned from the VQA training set. However, learning knowledge from image-question-answer triplets in the training dataset is not scalable and is liable to biases in the training datasets. We should exploit other external knowledge sources such as Wikipedia or knowledge graphs. Recently OK-VQA~\cite{marino19cvpr} provided a dataset consisting of these types of questions to let us better study open-domain knowledge in VQA.

We can define two types of knowledge representation that can be useful for these types of questions: First we have implicit knowledge, knowledge which is embedded into some non-symbolic form such as the weights of a neural network derived from annotated data
or large-scale unsupervised language training. Recently, Transformer- and specifically BERT~\cite{devlin19bert}-based multi-modal VQA models have been proposed~\cite{li2019visualbert, lu19vilbert, lu202012}, which incorporate large scale language pretraining, implicitly capturing language based knowledge. 
This type of knowledge can be quite useful, but we find this form of implicitly learned knowledge is not sufficient to answer many knowledge-based questions as we will show. 
Perhaps this is not surprising if one considers that many knowledge facts are very rare such as ``Thomas Newcomen invented the steam engine'' and learning them with hidden implicit representations might be less efficient while there are external sources and knowledge bases that state it explicitly. 

The other type of knowledge typically studied 
is explicit or symbolic knowledge, often in the form of
knowledge graphs. 
Approaches that use this form of knowledge either take the symbolic knowledge and then embed-and-fuse
them into a larger VQA model before answer prediction which no longer maintains the well-defined knowledge structures~\cite{marino19cvpr, guohao20mm}, or by relying on a closed set of knowledge facts with strong annotation of source knowledge~\cite{narasimhan2018out, wang17b, wu16}. In the second case, the VQA dataset itself has ground truth ``facts'' associated with the question, so solving these questions often ends up being the problem of retrieving a fact from the closed set.
In our method, we preserve the symbolic meaning of our knowledge from input until answer prediction. This allows us to use knowledge that is rare or is about rare entities as learning the reasoning logic with symbols is shared across all symbols.
And unlike other work, we do not have a closed set or ground truth knowledge, so we must build a large diverse knowledge base for use by our model.

In this work, we develop an architecture, \emph{\ModelName (\ModelNameLong)}, to successfully combine the implicit and symbolic knowledge. 
Specifically, \ModelName uses (i) a multi-modal BERT-pretrained transformer to process the question and image, and take advantage of the implicit knowledge in BERT, and (ii) a graph network to make use of symbolic knowledge bases. 
To cover the wide variety of knowledge required in OK-VQA, we draw on four very different knowledge sources to construct our knowledge graph: DBPedia~\cite{auer2007dbpedia}, ConceptNet~\cite{liu2004conceptnet}, VisualGenome~\cite{krishnavisualgenome} and hasPart KB~\cite{bhakthavatsalam2020dogs}.
This covers crowdsourced data, visual data, encyclopedic data, knowledge about everyday objects, knowledge about science and knowledge about specific people, places and events.
Finally, our method preserves the symbolic meaning of the knowledge by making predictions based on the hidden state of individual nodes in the knowledge graph
and using a late-fusion strategy to combine the implicit and symbolic parts of the model.

The main contributions of this work are as follows:
\begin{compactenum}
    \item We propose \ModelName (\ModelNameLong), a novel model incorporating explicit reasoning in a knowledge graph with implicit reasoning in a multi-modal transformer.
    \item Our model sets a new state-of-the-art on the challenging Open Knowledge VQA dataset (OK-VQA)~\cite{marino19cvpr}, significantly outperforming prior work.
    \item We extensively ablate our model to analyze various knowledge fusion strategies.
    \item We analyze how our model explicitly reasons about facts and answers questions by predicting answers from its knowledge graph. 
\end{compactenum}
\sectionvspace
\section{Related Work}
\sectionvspace
\myparagraph{Multimodal Vision and Language Modeling}
Approaches for multimodal vision and language tasks have explored diverse set of  fusion strategies such as bilinear models (\eg \cite{fukui2016multimodal,kim2018bilinear}) or self-attention (\eg \cite{gao2019dynamic}).
 Many recent works have been inspired by the success of Transformer \cite{vaswani2017attention} and BERT \cite{devlin19bert} models for natural language tasks and   proposed transformer-based fusion between image and text \cite{alberti2019fusion,chen2019uniter,li2019unicoder,li2019visualbert,lu19vilbert,su2019vl,tan2019lxmert,zhou2019unified}.
 Similar to these works
 as part of our method we train a multimodal transformer with BERT-pretraining to import the implicit language knowledge learned by BERT and learn any knowledge implicitly encoded in the training data and study how it fares on knowledge focused VQA.
 
 Another line of work for VQA has been extracting  programs from the question for more explicit reasoning with modules \cite{andreas16} or extracting symbols from the image to reason over them \cite{yi2018neural}. These works focus on reasoning about things explicitly shown in the image but do not integrate any external knowledge.
 
\myparagraph{Knowledge in Computer Vision}
Knowledge has a long history in computer vision problems. Some of the earliest versions of this work was relating to attributes~\cite{Farhadi09, Shrivastava14} or knowledge mined from the web \cite{rohrbach11cvpr}, often for zero- or few-shot learning problems~\cite{FeiFei06, Lampert14,rohrbach13nips}, as well as for fine-grained classification~\cite{Kun12}. The use of word embeddings from language has been extensive including in~\cite{Frome13, kottur2016visual,lu2016visual}. Class hierarchies such as WordNet~\cite{Miller95} have often been used to aid in image recognition~\cite{zhueccv14, redmon2017yolo9000}. Knowledge graphs have also found extensive use in visual classification and detection~\cite{marino17, chen2018iterative}, zero-shot classification~\cite{Wang_zslCVPR2018} and image retrieval~\cite{FeiFei15_2}. In our work we also rely on a knowledge graph to represent symbolic knowledge.

\myparagraph{Knowledge-based VQA datasets}
While open-ended VQA datasets (\eg \cite{antol15}) might require outside knowledge to answer some of its questions which cannot be learned from the dataset,  there are a few datasets which focus specifically on knowledge based multi-modal reasoning. One is FVQA \cite{wang17b}, where image-questions-answer triples are annotated with a fact-triple (\eg  “chair is furniture") from a fixed outside knowledge base, which allows deriving the answer. Specifically one of the two nodes (\ie chair or furniture in this example) is the answer. A more recent and more challenging dataset is OK-VQA \cite{marino19cvpr} which stands for \emph{Open Knowledge VQA}, as the name suggests, focusing on knowledge which is not tied to a specific knowledge base.

In this work we focus our evaluation on OK-VQA due to its relatively large number of knowledge-based questions, as well as its challenging and open-ended nature.

\myparagraph{Symbolic Knowledge for VQA}
Symbolic knowledge from knowledge bases is most commonly represented as graphs/knowledge bases~\cite{guohao20mm,narasimhan2018out, narasimhan18, wang17a, wang17b} or textual knowledge sources such as Wikipedia~\cite{marino19cvpr,wu16}. We can separate these works in two directions according to the criterion if the symbols are retained until answer prediction or not. \cite{narasimhan2018out,wang17a,wang17b} retain the symbols until the answers, allowing good generalization capabilities but require annotations of the ``correct" knowledge fact and are difficult to generalize to open knowledge VQA.
For improved generalization to open-domain VQA, \cite{garderes2020conceptbert,marino19cvpr,guohao20mm,wu16} embed the symbolic knowledge to an implicit embedding loosing the semantics of the symbols, but therefore are able to easily integrate the embedding with standard VQA approaches.  Similar to our work, the recent work \cite{garderes2020conceptbert} relies on a multimodal transformer model (pretrained VilBERT \cite{lu19vilbert}, 
however, similar to the other works it looses the semantics of the knowledge symbols when it integrates over them with an attention model.
In contrast, our work shows how to take advantage of both the implicit and symbolic knowledge directions: We retain symbols until the end without the need of knowledge-fact annotations and integrate it with implicit knowledge and powerful reasoning abilities of multi-modal transformers.

\myparagraph{Knowledge Bases \& Knowledge in NLP}
There have been a number of knowledge bases proposed and used for knowledge-based reasoning, both for language-only and multi-modal tasks~\cite{zhueccv14,neil,levan,sadeghi15,zhu15,ZhuLF17,bhakthavatsalam2020dogs, Miller95,krishnavisualgenome}. In the natural language processing literature, there has been much work in question answering from knowledge sources~\cite{berant13, yao14, bordes14} including for open-domain question answering~\cite{chen2017reading, wang2017r, yang2015wikiqa, yang2019end}.

\begin{figure*}[t]
\centering
\includegraphics[width=0.9\linewidth]{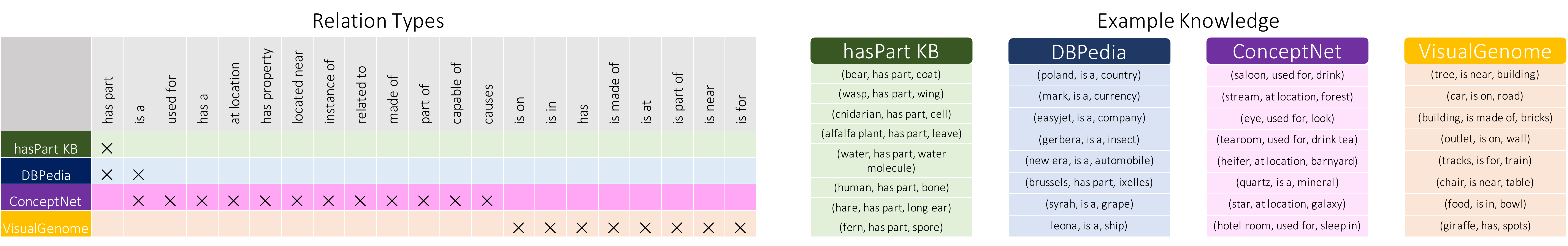}
\caption{Example knowledge and edge types from our knowledge graph from our four sources of explicit knowledge.}
\vspace{-.3cm}
\label{fig:knowledge}
\end{figure*} 

\sectionvspace
\section{The \ModelName Model}
\sectionvspace
In this section we introduce our model: \emph{\ModelNameLong} (\ModelName). An overview of our model can be seen in Fig.~\ref{fig:model}. We first introduce our transformer-based multi-modal implicit knowledge reasoning (Sec.~\ref{sec:model:implicit}), then discus the symbolic knowledge sources and reasoning with symbols (Sec.~\ref{sec:model:symbol}), and then describe their integration in Sec.~\ref{sec:model:integration}.

\sectionvspace
\subsection{Reasoning with Implicit Knowledge}
\label{sec:model:implicit}
\sectionvspace

\label{sec:VB}
We want to incorporate implicit external knowledge as well as multi-modal knowledge which can be learned from training set in our model.
Language models, and especially transformer-based language models, have shown to contain common sense and factual knowledge \cite{petroni19emnlp, jiang2020can}.
Most recent multi-modal models have also relied on the transformer architecture to learn vision-and-language alignment \cite{li2019visualbert,lu19vilbert}. 
We adopt this direction in our work and build a multi-modal transformer model, pre-trained with BERT~\cite{devlin19bert}, which has been pre-trained on the following language corpora to capture implicit knowledge: BooksCorpus~\cite{zhu15aligning} (800M words) and English Wikipedia~\cite{wikipedia} (2.5B words).
To learn multi-modal knowledge from the training set, our model is most closely related to the architecture used in \cite{li2019visualbert}. We also explore multi-modal pre-training in Section \ref{sec:soa}.

\myparagraph{Question Encoding}
We tokenize a question $Q$ using WordPiece~\cite{wu2016google} as in BERT~\cite{devlin19bert}, giving us a sequence of $|Q|$ tokens and embed them with the pre-tained BERT embeddings and append BERT's positional encoding, giving us a sequence of $d$-dimensional token representation $x_1^Q,...,x_{|Q|}^Q$. We feed these into the transformer, finetuning the representation during training.

\myparagraph{Visual Features}
As with most VQA systems, we use visual features extracted on the dataset by a visual recognition system trained on other tasks.
We use bottom-up features~\cite{anderson2018bottom} collected from the classification head of a detection model, specifically Faster R-CNN~\cite{renNIPS15fasterrcnn}. Because of the overlap in OK-VQA test and VisualGenome/COCO~\cite{LinMBHPRDZ14} trainval, we trained our detection model from scratch on VisualGenome, using a new split of VisualGenome not containing OK-VQA test images. The detector uses feature pyramid networks~\cite{lin2017feature}, and is trained using the hyper-parameters used for the baselines in~\cite{jiang2020defense}. 

We input bounding box features extracted from the image as well as the question words to the transformer. We mean-pool the output of all transformer steps to get our combined implicit knowledge representation $z^{implicit}$. 

\sectionvspace
\subsection{Reasoning with Symbolic Knowledge}
\label{sec:model:symbol}
\sectionvspace

\myparagraph{Visual Symbols}
In addition to using a pre-trained visual recognition system to get image features,
we also extract visual concepts (i.e. the predictions). This not only allows us to get a set of concepts to use to prune our knowledge graph (see Sec.~\ref{sec:knowledgegraphs}), it also gives us an entry point to get from the raw image to a set of symbols. This is significant---in order for our graph network to be able to reason about the question, it not only needs to reason about the question itself, but the entities in the image. For instance, if a question were to ask ``what is a female one of these called?'' in order use our knowledge that a female sheep is called an ``ewe,'' the graph network needs to actually know that the thing in the picture is a sheep. As we will see, being able to use these symbols is critical for our graph network to reason about the question.

There are a number of visual concepts we want to cover: places, objects, parts of objects and attributes. Therefore we run four classifiers and detectors trained on images from the following datasets: ImageNet~\cite{ILSVRC15} for objects, Places365~\cite{zhou2017places} for places, LVIS~\cite{gupta2019lvis} for objects and object parts and Visual Genome~\cite{krishnavisualgenome} for objects, parts and attributes. This gives us a total of about 4000 visual concepts. We give additional details about these classifiers in Appendix~\ref{appx:imagesymbols}.

\myparagraph{Knowledge Graph Construction}
\label{sec:knowledgegraphs}

Unlike previous work such as~\cite{narasimhan2018out}, or in NLP work on datasets such as SQuAD  \cite{rajpurkar16squad} which study the problem of closed-system knowledge retrieval, we do not have a ground truth set of facts or knowledge which can be used to answer the question. We must make an additional choice of what knowledge sources to use and how to clean or filter them.

There are a few different kinds of knowledge that might help us on this task. One is what you might call trivia knowledge: facts about famous people, places or events. Another is commonsense knowledge: what are houses made of, what is a wheel part of. Another is scientific knowledge: what genus are dogs, what are different kinds of nutrients. Finally, situational knowledge: where do cars tend to be located, what tends to be inside bowls. 

The first and largest source of knowledge we use is DBPedia~\cite{auer2007dbpedia}, containing millions of knowledge triplets in its raw form. DBPedia is created automatically from data from Wikipedia~\cite{wikipedia}. This tends to give a lot of categorical information e.g. (Denmark, is\_a, country), especially about proper nouns such as places, people, companies, films etc. The second source of knowledge is ConceptNet~\cite{liu2004conceptnet}, a crowd-sourced project containing over 100,000 facts organized as knowledge triples collected  by translating English-language facts into an organized triplet structure. It also contains as a subset the WordNet~\cite{Miller95} ontology. This dataset contains commonsense knowledge about the world such as (dog, has\_property, friendly). Following~\cite{marino17}, we also use the scene graphs from VisualGenome~\cite{krishnavisualgenome} as another source of knowledge. As in~\cite{marino17}, we take a split of VisualGenome that does not contain any OK-VQA test images. This knowledge source tends to give us more spatial relationships e.g. (boat, is\_on, water) and common pairwise affordances e.g. (person, sits\_on, coach). Finally, we use the new hasPart KB~\cite{bhakthavatsalam2020dogs} to get part relationships between common objects such as (dog, has\_part, whiskers) as well as scientific ones (molecules, has\_part, atoms). We show example knowledge triplets from our in Fig.~\ref{fig:knowledge}. 

With these knowledge sources, we can capture a large amount of knowledge about the world. But we then run into a problem of scale. In its raw form, DBPedia alone contains millions of edges, with the others containing a total of over 200,000 knowledge triplets. This first presents a technical problem---this graph is far too large to fit into GPU memory if we use a graph neural network model. But more fundamentally, while this knowledge graph contains a lot of useful information for our downstream task, it also includes a lot of irrelevant knowledge. In particular, DBPedia, being parsed automatically from Wikipedia pages, contains information about virtually every film, book, song and notable human in history. While some of those may be useful for particular questions, the vast majority is not.

To deal with these issues, we limit our knowledge graph to entities that are likely to be helpful for our end task. First, we collect all of the symbolic entities from the dataset: in particular the question, answers and visual concepts that can be picked up by visual recognition systems (see Sec.~\ref{sec:model:symbol}). We then include edges that only include these concepts

\footnote{As before, we use the training set to avoid data leakage.}. After this filtering, we have 
a total of about 36,000 edges and 8,000 nodes. We provide more exhaustive details of our knowledge collection and filtering in Appendix~\ref{appx:kgconstruction}.

\begin{figure*}[t]
\centering
\includegraphics[width=0.85\linewidth]{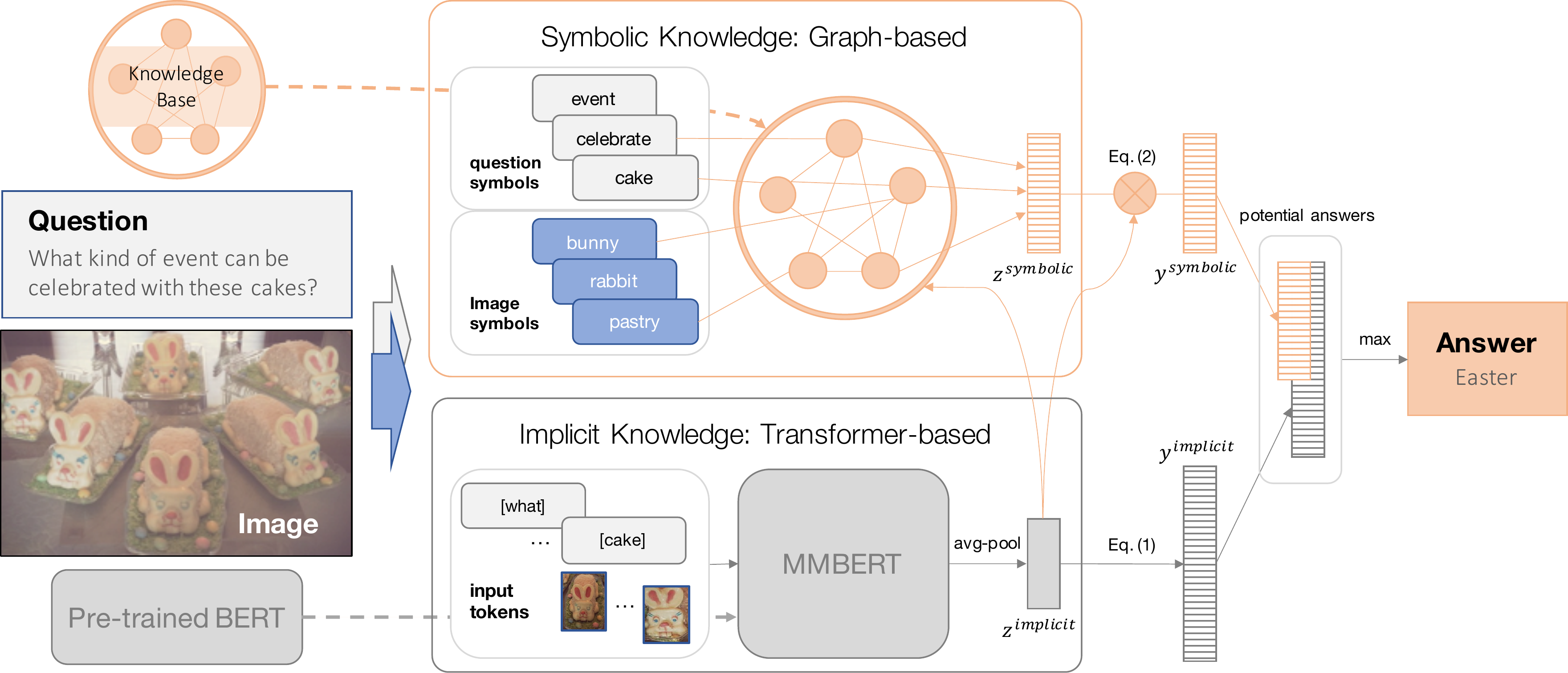}
\vspace{.1cm}
\caption{Our model \ModelName integrates implicit knowledge and reasoning (bottom) with explicit graph-based reasoning on a knowledge base (top). The implicit knowledge model receives the visual features and question encoding whereas the explicit knowledge model operates on image and question symbols. They predict answers according to Eq.~\ref{eq:implicit}\&\ref{eq:symbolic} and we take the max overall prediction (see Sec.~\ref{sec:model:integration}).
}
\vspace{-.3cm}
\label{fig:model}
\end{figure*}

\myparagraph{Graph Network}
\label{sec:graphnetwork}
Now we move to our symbolic knowledge representation. We want to treat our knowledge graph as input without having to decide on which few facts out of our entire knowledge graph might be relevant. So to process on our entire knowledge graph and decide this during training, we use a graph neural network to incorporate our knowledge.

In our network, each node of the graph network corresponds to one specific symbol representing one concept such as ``dog'' or ``human'' in our knowledge graph.

The idea is that the graph neural network can take in information about each specific symbol and use the knowledge edges to infer information about other symbols by passing information along the edges in the knowledge graph. And, in our  
graph neural network we share the network parameters across all symbols, meaning that unlike for other types of networks, the reasoning logic is shared across all symbols which should allow it to generalize better to rare symbols or graph edges.

We use the Relational Graph Convolutional Network (RGCN)~\cite{schlichtkrull2018modeling} as the base graph network for our model. Unlike the related GCN~\cite{kipf2016semi}, this model natively supports having different calculations between nodes for different edge types (an is\_a relationship is treated differently than a has\_a relationship) and edge directions (dog is\_a animal is different than animal is\_a dog). With this architecture we also avoid the large asymptotic runtime of other architectures with these properties such as~\cite{Li16} or~\cite{velivckovic2017graph}.

\myparagraph{Graph Inputs} 
For one particular question image pair, each node in the graph network receives $4$ inputs. 
1. An indicator $0/1$ of whether the concept appears in the question.
2. The classifier probabilities from Sec.~\ref{sec:model:symbol} for the node's concept, (or $0$ if the concept is not detected in the particular image) 
With $4$ image classifiers or detectors, the node receives $4$ separate numbers. 3. The $300d$ word2vec (GloVe~\cite{pennington2014glove}) representation of that concept, or average word2vec for multi-word concepts. 4. The implicit knowledge representation $z^{implicit}$ from Sec.~\ref{sec:VB} passed through a fully connected layer: $fc(z^{implicit})$ with ReLU activation to reduce the size of this feature to $128$ for efficient graph computation.

Following the standard formulation of graph neural networks, we write the input to the graph neural networks (described above) as $X = H^{(0)}$ where $X$ is a $\mathbb{R}^{n \times d_s}$ matrix with $n$ node inputs of size $d_s=433$.
Then for each layer of the RGCN, we have a non-linear function $H^{(l+1)} = f(H^{(l)}, KG)$ where $KG$ is the knowledge graph. The RGCN convolution uses different weight matrices for different edge types and for different directions. As a result the semantic difference between an is-a relationship and a has-a relationship as well as the direction of those edges is captured in the structure of the network and different transformations are learned for each. After all RGCN layers are computed we end up with $H^{(L)} = G$ which is a $\mathbb{R}^{n \times d_h}$ matrix which corresponds to having a hidden state of size $f_h$ for each node (and therefore concept) in our graph.

Additional architectural details and parameters of the graph network can be found in Appendix~\ref{appx:hyper}.

\sectionvspace
\subsection{Integrating Implicit and Symbolic Knowledge}
\label{sec:model:integration}
\sectionvspace

Finally, given the output of our implicit transformer-based module $z^{implicit}$ and our explicit/symbolic module $G$, how do we get our final prediction? Our main insight to make a separate prediction for $z^{implicit}$ and for each node/concept in the knowledge graph.

\myparagraph{Implicit Answer Prediction}
As is now commonplace among VQA methods, to get the implicit answer prediction, we do a final prediction layer and predict the answer within a set vocabulary of answers $V \in \mathbb{R}^{a}$ where $a$ is the size of the answer vocabulary. We simply have:
\begin{equation}
y^{implicit}{=}\sigma(W z^{implicit}{+}b)
\label{eq:implicit}
\end{equation} 
where $\sigma$ is the sigmoid activation.

\myparagraph{Symbolic Answer Prediction}
To predict the answers for symbolic, we note that $G$ can be rewritten as a hidden state node $z_{i}^{symbolic}$ for each node/concept $i$ in the knowledge graph. Because each of these nodes corresponds to a word or multi-word symbol, we actually have nodes and corresponding hidden states that are possible answers to a $VQA$ question. So for each hidden state that is in our answer vocab $V \in \mathbb{R}^{a}$ we make a prediction for it.

For each of these answer nodes $i$, we predict:
\begin{equation}
\resizebox{0.9\hsize}{!}{
        $y_i^{symbolic}{=}\sigma((W^s z_i^{symbolic}{+}b^s)^T(W^z z^{implicit}{+}b^z)).$      
        }
\vspace{-.05cm}
\label{eq:symbolic}
\end{equation}

We additionally re-use the implicit hidden state $z^{implicit}$ to make this prediction. This gives us an additional late fusion between the implicit and symbolic parts of our model.

\myparagraph{Final Prediction}
Finally, given our final predictions from each part of the network $y^{implicit}$ and $y^{symbolic}$,we  can simply  choose  the  final  answer  by  choosing  the  highest scoring answer from both  answer vectors. For training, we can simply optimize $y^{implicit}$ and $y^{symbolic}$ separately with a binary cross entropy loss end-to-end through the entire network. See Fig.~\ref{fig:model}.

\sectionvspace
\section{Results}
\sectionvspace
\subsection{Experimental Setup}
\label{sec:setup}
\sectionvspace
For all experiments, we train our models with  PyTorch~\cite{paszke2019pytorch} and the MMF Multimodal Framework~\cite{singh2020mmf}. We use PyTorch Geometric~\cite{fey2019fast} for our graph neural network implementations. We use the default training hyperparameters from MMF which we provide in Appendix~\ref{appx:hyper}. For consistency, for each result we train each model on $3$ random seeds and take the average as the result. We show sample std on these runs in Appendix~\ref{appx:variance}. 

For the purpose of state-of-the art comparisons in Table~\ref{table:OKVQA}, we compare our main method on the 1.0 version of OK-VQA~\cite{marino19cvpr}. Recently, a 1.1 version of the dataset was released, and all other experiments including ablations are done on this version. The only change between the versions is a change in how answer stemming is handled, resulting in a more coherent answer vocabulary. In particular, we observe that the new answer vocab has much fewer ``non-word'' stemming such as ``buse'' for busses and ``poni tail'' instead of ``pony tail.'' Unless otherwise stated, an experiment is on version 1.1.

For many of our ablations and analysis we train just the \MMBERTLong (\MMBERTAbr) model described in Sec.~\ref{sec:VB} by itself. Unless otherwise stated, this model and ours is always initialized from BERT.

In Sec.~\ref{sec:modelanalysis} we do a through ablation of \ModelName comparing the different parts of the model and design choices we made. In Sec.~\ref{sec:quantanalysis} we show the results of a number of experiments to more thoroughly analyze our method, especially looking at its performance on rare answers. In Sec.~\ref{sec:qualitative} we look at some specific questions and predictions from our model to get a more grounded idea of what our model does on real examples. Finally, in Sec.~\ref{sec:soa} we add visual-linguistic pre-training to our models to show how our model achieves state-of-the-art performance on OK-VQA.

\sectionvspace
\subsection{Model Analysis and Ablations}
\label{sec:modelanalysis}
\sectionvspace
We first analyse our model to see where the improvement is coming from with several ablations, especially focusing on symbolic vs. implicit knowledge and their integration. We want to understand which parts of our method are working and why.

\myparagraph{Ablation of Symbolic Knowledge}
First, we see how much of improvement comes from the \MMBERTBase backbone of our model versus from the symbolic Graph Network. In Table~\ref{table:AllAblation} (lines 1\&2),
we see that \ModelName combining implicit and symbolic knowledge improves significantly over the \MMBERTBase~by about 3\%. 

We should, however, make sure this improvement is due to the symbolic knowledge and not merely from a more complex or better architecture. While our \ModelName only has slightly more parameters (116M parameters versus \MMBERTAbr with 113M), it does add at least some extra computation. To test this, we approximate a version of our method with only  the architecture and not the underlying knowledge. To do this, we keep all network details the same, but instead of using the knowledge graph we constructed in Sec.~\ref{sec:knowledgegraphs}, we use a randomly connected graph. We keep all of the nodes the same, but we randomize the edges connecting them. So in this version with a random graph, our graph network receives all of the same inputs and the outputs, but all connections are completely random. If the performance were just from the computation, we would expect this to work. 
Instead, we see from line 3 that the performance using the random graph drops significantly.

\begin{table}[t]
\begin{center}
\begin{tabular}{@{}rlc@{}}
\toprule
&Method & accuracy\\ \midrule
1. & \ModelName (ours) & \bf{32.31} \\
\hline
\multicolumn{3}{c}{\textbf{Ablation of Symbolic Knowledge}}\\
2. & \MMBERTAbr & 29.26 \\
3. & \ModelName w/ random graph & 30.15 \\
\hline
\multicolumn{3}{c}{\textbf{Ablation of Implicit Knowledge}}\\
4. & \ModelName  w/o BERT pretrain & 26.28  \\
5. & \MMBERTAbr w/o BERT pretrain & 21.82  \\
\hline
\multicolumn{3}{c}{\textbf{Ablation of Network Architecture}}\\
6. & \ModelName no late fusion & 31.10 \\
7. & \ModelName no \MMBERTAbr input & 31.10 \\
8. & \ModelName no \MMBERTAbr input or late fusion & 25.00 \\
9. & \ModelName no backprop into \MMBERTAbr & 27.98 \\
10. & \ModelName with GCN & 30.58 \\
11. & \ModelName feed graph into \MMBERTAbr & 30.99 \\
\hline
\multicolumn{3}{c}{\textbf{Ablation of Graph Inputs}}\\
12. & \ModelName no Q to graph & 31.74\\
13. & \ModelName no I to graph & 31.59\\
14. & \ModelName no symbol input & 30.26 \\
15. & \ModelName no w2v & 31.95  \\
\bottomrule
\end{tabular}
\end{center}
\caption{\ModelName ablation on OK-VQA v1.1. We show the performance of our model compared with the implicit-only baseline (\MMBERTAbr). We also show ablations without BERT training, with a random knowledge graph, ablations on our model architecture, and ablations where we remove the question input to the graph network (no Q), the image inputs (no I) and both (no symbol).}
\label{table:AllAblation}
\vspace{-.2cm}
\end{table}

\myparagraph{Ablation of Implicit Knowledge}
Next we look at the implicit knowledge contained in the BERT versus our combined system 
to see how much of an effect it had. From Table~\ref{table:AllAblation} we can see that BERT is a crucial element. Without the BERT pre-training (lines 4\&5), our method falls by 6\% and the \MMBERTBase falls by an even larger 7\%. This shows that the implicit knowledge is an important component of our model. The difference between \ModelName and \MMBERTBase when neither has BERT pre-training is actually higher than the difference with BERT, about 4.5\%, suggesting that there is some overlap in the knowledge contained in our knowledge graphs with the implicit knowledge in BERT, but most of that knowledge is non-overlapping.

\myparagraph{Ablation of Network Architecture}
Next, we want to get a sense of which parts of our architecture were important. As we can see, our particular architecture is critical: the use of \MMBERTAbr features as input to \ModelName and the late fusion were both important.
With just one of these (lines 6\footnote{We replace Eq.~\ref{eq:symbolic} with a linear layer that only takes in $z_i^{symbolic}$.}\&7), performance drops by about 1\%, but without either (line 8), performance drops over 7\%. Without at least one connection between the \MMBERTBase and the graph network, there can be no fusion of the visual features and question and the graph network cannot incorporate any of the implicit knowledge in BERT. We also tried \ModelName where these two ways of fusing were present, but we did not allow any backpropogation from the Graph Network to \MMBERTAbr (line 9). This also performs badly, as the graph network cannot correct errors coming from this input, but not as bad as removing these connections entirely (line 8).

We also tried a less powerful graph network: GCN~\cite{kipf2016semi} (line 10) which critically does not have directed edges or edge types. This baseline hurts performance by about 2\% justifying our choice of a graph network that uses edge direction and type. We also have another architectural ablation, where we feed the graph network features directly to the \MMBERTBase rather than having a separate answer prediction directly from the graph as in \ModelName or any of the other baselines (line 11). This architecture performs much worse than our final model.

\myparagraph{Ablation of Graph Inputs}
Next we look at the symbolic and non-symbolic inputs to the knowledge graph nodes to see what effect those might have had in the next section of Table~\ref{table:AllAblation}. First, we ablate the question indicator input (line 12) and the image confidences (line 13) described in Sec.~\ref{sec:graphnetwork}. We find that removing one or the other drops performance, but not drastically, but dropping both (line 14) drops performance by about 2\%, much more than the effect of dropping the \MMBERTAbr input to the graph. We also ablate the word2vec inputs to nodes (line 15) and find that this part made the least difference, dropping it less than 1\%.

\begin{table}[t]
\begin{center}
\begin{tabular}{@{}rlc@{}}
\toprule
&Method &  accuracy\\ 
\hline
1. & \ModelName $\max(y^{implicit},y^{symbolic})$ (ours) & \bf{32.31} \\
2. & \ModelName $y^{implicit}$ & 31.47 \\
3. & \ModelName $y^{symbolic}$ & 29.36 \\
4. & \ModelName no backprop $y^{implicit}$ & 28.19\\
\hline
5. & \ModelName $\text{oracle}(y^{implicit}|y^{symbolic})$  & 36.71\\
\bottomrule
\end{tabular}
\end{center}
\caption{\ModelName Subpart Analysis on OK-VQA v1.1. Here we show the OK-VQA accuracy of different parts of the model separately: just the \MMBERTAbr ($y^{implicit}$), just the graph network ($y^{symbolic})$. We also show the \MMBERTAbr only without a backpropogation signal between the two parts and an oracle best-case performance between the two parts.}
\vspace{-.2cm}
\label{table:GraphAnalysis}
\label{sec:analysis}
\end{table}

\myparagraph{Preserving Symbolic Meaning}
One major claim we make is that symbolic and implicit knowledge are both necessary for this problem. The results without BERT training make the case pretty clearly that implicit, non-symbolic knowledge from BERT is critical. From the ablation of symbolic knowledge, we show that it is the symbolic knowledge (and not just the architecture) greatly contributes to the performance of our method.
On the symbol input side, we show that removing the symbolic inputs (line 12) 
hurts performance, even more than removing the \MMBERTBase hidden input (line 7) which  contains information about the same image and question, but in a non-symbolic form. 
Finally we have a baseline (line 11) where instead of predicting separate outputs from the graph network and \MMBERTBase, we directly connect the graph network into \MMBERTAbr, feeding a pooled graph hidden state (see Appendix~\ref{appx:gntovbbaseline} for more details)
into the \MMBERTAbr 
as an input. This baseline does significantly worse. What these ablation have in common is that they remove the direct connection between the symbols and the knowledge graph. When the graph network is not able to connect the knowledge symbolically to the input symbols or the output symbols, we see that it performs worse. In addition, we know symbolic knowledge itself is useful because when we only change the connections between nodes and nothing else (line 3), performance drops drastically. Our entire graph module directly connects symbols in the input (question words and image symbols from classifiers) to symbols that are the output (the answer words) and this seems to be critical to the performance of \ModelName. 
\sectionvspace
\subsection{Quantitative Result Analysis}
\label{sec:quantanalysis}
\sectionvspace
First we examine the parts of our model separately to see if we can learn anything about how the \MMBERTAbr and Graph Network parts of \ModelName interact. 

\begin{table}[t]
\begin{center}
\begin{tabular}{@{}lcccc@{}}
\toprule
\multicolumn{1}{r}{Metric$\rightarrow$}&\multicolumn{2}{c}{Frequency Rank}&\multicolumn{2}{c}{\# Unique answers}\\
Method $\downarrow$ & All & Correct & All  & Correct \\ \midrule
\ModelName (ours) & \textbf{528.5} & \textbf{456.7} & \textbf{1349} & \textbf{780} \\
\MMBERTAbr  & 467.1 & 427.4 & 1247 & 719 \\
\bottomrule
\end{tabular}
\end{center}
\caption{Long-tail Analysis. We show \ModelName and the non-symbolic  \MMBERTAbr long-tail metrics for ``all'' predictions made by the model and for ``correct'' predictions. Higher is better.}
\label{table:LongTail}
\vspace{-.2cm}
\end{table}

\begin{figure*}[t]
\centering
\includegraphics[width=0.88\linewidth]{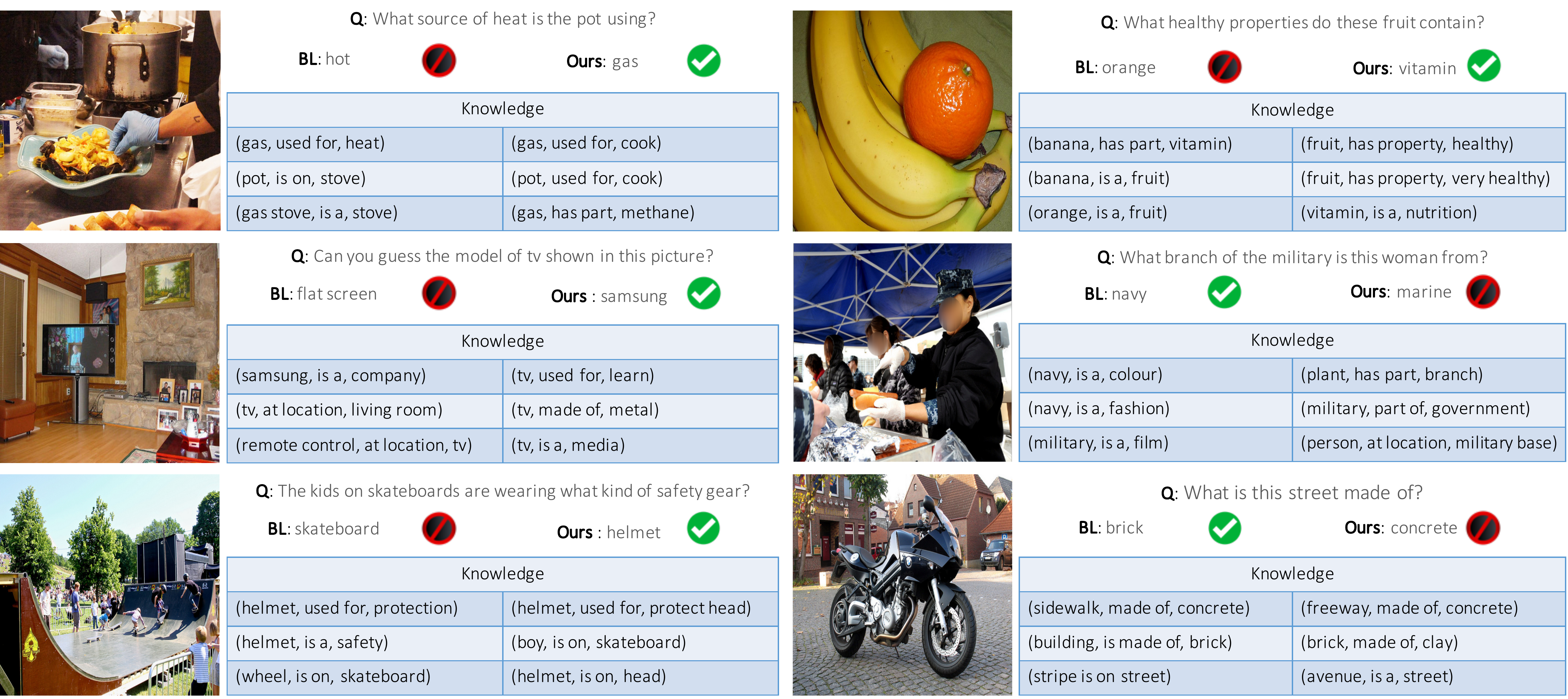}
\caption{Qualitative examples from \ModelName. Showing predictions by our model and the implicit knowledge baseline \MMBERTBase. We show the question, image, and answers given by both models. We also show knowledge in the graph related to the question, answers or image that seemed most relevant.}
\label{fig:qualitative}
\vspace{-.2cm}
\end{figure*}

In Table~\ref{table:GraphAnalysis} we look at the performance of different parts of our model (without retraining the model for lines 1,2,3,5). Since the \MMBERTAbr~and Graph Network parts of \ModelName~produce separate predictions, we can analyze them separately. 
For instance, we find that despite the fact that the \MMBERTAbr~part of our model does not receive input from the Graph Network, the \MMBERTAbr~(Table~\ref{table:GraphAnalysis}, line 2) has a higher accuracy of 31.47\% than the \MMBERTAbr~baseline (Table~\ref{table:AllAblation}, line 2), 29.26\%. This we suspect is because this part of the network receives a back-propagation from the Graph Network part of the model and this extra component improves the quality of the \MMBERTAbr~pooled feature because it is also trained to reduce the loss from the late fusion predictions. Indeed, if we remove the back-propagation signal  
(Table~\ref{table:GraphAnalysis}, line 4) we see that the accuracy of this part of the model drops down to 28.19\%. We also see a direct improvement beyond this effect. Comparing the \MMBERTBase (line 2) and Graph Network (line 3) -only accuracies, the Graph Network does a bit worse on its own, but not by a huge amount, and the Graph Network predictions are used 47\% of the time in the joint model (line 1). Since the accuracy of the combined model is higher than each, it is able to choose the correct answer from between \MMBERTAbr and Graph Network. Finally, we see that if we had an oracle that always chose the best prediction from either the \MMBERTAbr or the Graph Network, we would improve the accuracy to 36.71\%! Obviously this is not a realistic number to achieve since it uses ground truth, but it shows that the \MMBERTAbr and Graph Network predictions are non-redundant.

\myparagraph{Long-Tail Analysis}
Next, we try to see whether our explicit/implicit model performs any differently on the ``long tail'' of OK-VQA. OK-VQA itself is built as a long-tail dataset, specifically rejecting answers that appear too many times to avoid models overfitting to the answer vocabulary, making it a good dataset to study knowledge-based VQA. Even with this filtering, some answers do appear more often than others, so we can try to study whether our method does better on rare answers. 

In Table~\ref{table:LongTail} we show metrics on \ModelName versus the baseline \MMBERTBase. First we use a metric we refer to as ``Answer Frequency Rank''. This simply means we order the answers in the dataset from most common to least common 
and assign them a rank from 1 for the most common to the total number of answers in the dataset. On this metric our model 
scores higher, which means it chooses on average less common answers. This is true whether you measure for all prediction or for only correct predictions.
For a perhaps more intuitive metric we also look at the number of unique answers our model predicts versus the baseline. Here we predict $1349$ versus $1247$ or $780$ versus $719$ if we only look at correct predictions.
These results indicate that our model is generalizing better to the long-tail.
\sectionvspace
\subsection{Qualitative Analysis}
\label{sec:qualitative}
\sectionvspace
Finally, we show some examples of our model to how our knowledge graph might be helping answer questions. It is obviously good to analyze our method quantitatively to get an objective sense of what our method is doing, but seeing what our model does on specific examples can be very instructive. See Fig.~\ref{fig:qualitative}. In the top left example we see an example where our model correctly guesses that the source of heat for the pot is ``gas.'' 
Looking at the knowledge graph, some knowledge that may have been helpful was that gas is used for heat, that both gas and pot are used to cook. The knowledge graph here connects directly from a word in the question to the answer. In the next question, it asks what model the tv is and it guesses Samsung. This is supported by an edge that indicates that Samsung is a company which makes it more likely to be a ``model'' of a product. 
In the next example we see that there is knowledge connecting the answer ``helmet'' to the word ``safety'' in the question as well as other information such that helmet is used for protection that supports the answer. 
In the next question we see that we have knowledge that connects entities in the question to the image (both oranges and bananas are fruit) and information that bananas have vitamins, connecting directly to the answer ``vitamin.''
Finally, we show some examples our method did poorly on. For the next question, our method says the woman in the picture is a marine instead of navy. Here it may have been confused by knowledge that navy is a color or a fashion, showing that homonyms can be a problem for symbolic knowledge. Finally for the last question, our method answers concrete instead of brick, likely because the available knowledge supported this. 
We include more examples in Appendix~\ref{appx:qual}.

\sectionvspace
\subsection{Pre-training and State-of-the-Art Comparison}
\label{sec:soa}
\sectionvspace
In this section we study the benefit of visio-linguistic pre-training which has shown to be beneficial for many vision-and-language tasks (see e.g. \cite{lu19vilbert,li2019visualbert}) including OK-VQA \cite{garderes2020conceptbert} and compare the results to prior work.

\myparagraph{Pre-training}
First, we look at three kinds of pre-training for our model and how it affects the performance.

The first two are Masked COCO and Masked VQA, introduced in ~\cite{singh2020we}. The objective is that given the image regions as $v = \{v_1, ..., v_N \}$, the input texts as $l = \{l_1, ..., l_M\}$ we train a model to reconstruct either $l$ and/or $v$ from corrupted versions $\hat{v}$ and $\hat{l}$ where some words
$l_m$ or image regions $v_n$ are masked. In the Masked COCO task, the captains are used as $l$ and for the Masked VQA task, the questions are used as $l$. The third task is simply training on the question answering objective of VQAv2~\cite{goyal2017making}.

\begin{table}[t]
\begin{center}
\begin{tabular}{@{}lcc@{}}
\toprule
Pre-training & \MMBERTAbr & \ModelName \\ \midrule
BERT only & 29.29  & 32.31 \\
Masked COCO Captions  & 33.19 & 35.04\\
Masked VQA Questions  & 34.32  & 35.74\\
VQAv2 & 37.10  & 37.79\\
VQAv2 (incl. graph)  & -- & \textbf{38.90}\\
\bottomrule
\end{tabular}
\end{center}
\caption{Vision/Language Pre-Training results on OK-VQA v1.1. We compare \MMBERTAbr and \ModelName on our three pre-training tasks, Masked COCO, Masked VQA and VQA. For the \MMBERTAbr model we only pre-train the transformer except in the last experiment where we pre-train the entire model (incl. graph).}
\label{table:Pretrain}
\vspace{-.3cm}
\end{table}

In Table~\ref{table:Pretrain} we show the results of \ModelName as well as the baseline \MMBERTAbr pre-trained on these tasks. Note that the transformers are still pre-trained on BERT---we do this pre-training starting from BERT. For all but the last line in the table, we only pre-train the transformer model on these tasks. For the final number, we pre-train our entire \ModelName model including the graph network on the VQA task.

As we can see, all forms of pre-training improve our models. The most effective method of pre-training is to train on VQA. This is intuitive since OK-VQA and VQA are quite similar tasks. We also see that our \ModelName model consistently outperforms \MMBERTAbr, which is our model without symbolic knowledge. 
Interestingly, we find that it is not only beneficial to pre-train the transformer but also the symbolic graph network (note that for \MMBERTAbr the entire model is pre-trained already in the second to last line as it does not have a graph component). Our fully pre-trained \ModelName achieves 38.90\% accuracy, compared to fully pre-trained \MMBERTAbr of $37.10$\%.

\myparagraph{OK-VQA v1.0 Comparison}
In Table~\ref{table:Pretrain} we set the bar for performance on OK-VQA v1.1 with an accuracy of $38.90$\%. In order to compare to other works (all of which show results on v1.0), we compute the performance of our best model (VQA joint graph and transformer pre-training) on OK-VQA v1.0 as well. We see that our model achieves $38.35$\% accuracies versus the best previous state-state-of-the-art of $33.66$\%~\cite{garderes2020conceptbert}.

We note from the last lines of Table~\ref{table:Pretrain} versus the last line in Table \ref{table:OKVQA} that our method performs a bit better on version 1.1 of the dataset, suggesting that indeed this is a cleaner handling of answer vocab.

\begin{table}[t]
\begin{center}
\begin{tabular}{@{}lc@{}}
\toprule
Method & accuracy\\ \midrule
Q-Only & 14.93 \\
MLP & 20.67  \\
BAN \cite{kim2018bilinear} & 25.17 \\
BAN+AN \cite{marino19cvpr} & 25.61 \\
BAN+KG-Aug \cite{guohao20mm} & 26.71 \\
MUTAN \cite{ben2017mutan} & 26.41 \\
MUTAN+AN \cite{marino19cvpr} & 27.84  \\
ConceptBERT \cite{garderes2020conceptbert} & 33.66 \\
\ModelName (ours) & \textbf{38.35} \\
\bottomrule
\end{tabular}
\end{center}
\caption{Benchmark results on OK-VQA v1.0}
\label{table:OKVQA}
\vspace{-.3cm}
\end{table}

\sectionvspace
\section{Conclusion}
\sectionvspace
In this paper we introduce \emph{\ModelNameLong} (\ModelName): a method for incorporating implicit and symbolic knowledge into Knowledge-Based VQA. We show it outperforms prior works on OK-VQA~\cite{marino19cvpr}, the largest available open-domain knowledge VQA dataset. We show through extensive ablations that our particular architecture outperforms baselines and other alternatives by preserving the symbolic representations from input to prediction. Moreover, through experiments, analysis, and examples we find our model makes use of both implicit and symbolic knowledge to answer knowledge-based questions and generalizes to rare answers.

\vspace{.2cm}

\noindent {\footnotesize {\bf Acknowledgements}: We want to thank Amanpreet Singh, Abhishek Das, Ronghang Hu and Vedanuj Goswami who provided data and help with the MMF Multimodal Framework and PyTorch Geometric. As part of their affiliation with CMU, Kenneth Marino is supported by the Department of Defense (DoD) through the National Defense Science \& Engineering Graduate Fellowship (NDSEG) Program. The Georgia Tech effort was supported in part by NSF, AFRL, DARPA, and ONR YIP. The views and conclusions contained herein are those of the authors and should not be interpreted as necessarily representing the official policies or endorsements, either expressed or implied, of the U.S. Government, or any sponsor.}

{\small
\bibliographystyle{ieee_fullname}
\bibliography{egbib}
}

\clearpage
\appendix

\section{Methodology Additional Details}
\label{appx:method}

\subsection{Knowledge Graph Construction}
\label{appx:kgconstruction}
Here we provide additional details on our knowledge graph construction.

\myparagraph{DBPedia}
First we extracted DBPedia~\cite{auer2007dbpedia}. DBPedia is actually a set of datasets collected from Wikipedia articles and tables. For our knowledge graph we used the October 2016 crawl of Wikipedia.\footnote{\url{https://wiki.dbpedia.org/downloads-2016-10}} For our DBPedia edges we used the following files: Article Categories, Category Labels, DBPedia Ontology, Instance Types, Instance Types Sdtyped Dbo, Mappingbased Objects, and Person Data. 

Next we wrote string parsers and regular expressions to translate these triplets into lowercase multi-word english expressions. This involved extracting the category words from the hyperlink: e.g., ``\texttt{\footnotesize <http://dbpedia.org/resource/Tadeusz\_Borowski>}'' would be extracted as ``tadeusz borowski''. 

Before final filtering, this knowledge source contains $24,685,703$ edges.

\myparagraph{VisualGenome}
As we say in the main text we collect a knowledge graph on VisualGenome~\cite{krishnavisualgenome} by taking the most common edges in the scene graphs. We first create a split of VisualGenome. So that this graph is maximally useful down the road, we take a split that only contains the intersection of COCO~\cite{LinMBHPRDZ14} train, VisualGenome train, and LVIS~\cite{gupta2019lvis} train so that the graph can safely be used for any of these datasets on COCO. This also means that this split does not contain and of OK-VQA~\cite{marino19cvpr} test set images.

For the remaining images, we take any edge which appear at least $50$ times in that set and add to our list.

Before final filtering, this knowledge source contains $3,326$ edges.

\myparagraph{hasPart KB / ConceptNet}
These two knowledge sources were already in a fairly processed state, so no additional processing was necessary before our task-specific filtering. hasPart KB~\cite{bhakthavatsalam2020dogs} was directly downloaded from source website.

ConceptNet~\cite{liu2004conceptnet} was from the training data used for~\cite{li2016commonsense} which has already been processed.\footnote{\url{https://ttic.uchicago.edu/~kgimpel/resources.html}}

hasPart KB contained $49,848$ edges and ConceptNet contained $102,400$.

\myparagraph{Combining and Filtering}
To combine and filter these four knowledge bases into one graph, the first step was to simply combine all of the knowledge triplets from the four knowledge sources. Then, we removed all stop word concepts (e.g. is, the, a) from the knowledge graph to avoid non-meaningful edges.

Next, as we discuss in the main text we collect all of the symbolic entities from the dataset (question, answers and visual concepts) and then include edges that only include these concepts. We also limit the number of $25$ edge types that are the most common and useful for our end task, shown in Fig.2 of the main text.

The final graph is $361,999$ edges, $7643$ nodes and $25$ edge types.

\subsection{Image Symbols}
\label{appx:imagesymbols}
To get our image symbols, as we say in the main paper, we run four classifiers and detectors on our dataset. The classifiers/detectors we use are the following.

\begin{enumerate}
    \item A ResNet-152~\cite{he2016deep} trained on ImageNet~\cite{ILSVRC15}. Implementation from default PyTorch~\cite{paszke2019pytorch} nn library.
    \item A ResNet-18 trained on Places365~\cite{zhou2017places} using that publication's released code. 
    \item A Faster R-CNN trained on Visual Genome~\cite{krishnavisualgenome} using the baseline from~\cite{jiang2020defense}.
    \item An EQL loss~\cite{tan2020equalization} -trained Mask R-CNN model on LVIS (v1.0)~\cite{gupta2019lvis} using the code from~\cite{tan2020equalization}.
\end{enumerate}

\begin{table}[h]
\begin{center}
\begin{tabular}{@{}lc@{}}
\toprule
Dataset & \# Symbols\\ \midrule
ImageNet& $1000$ \\
VisualGenome & $1600$ \\
LVIS & $1203$ \\
Places & $365$ \\
\bottomrule
\end{tabular}
\end{center}
\caption{\MMBERTBase Hyperparameters}
\label{table:datasetsize}
\end{table}

In Table~\ref{table:datasetsize} we show the number of symbols in each of these datasets.

\subsection{Answer vocab}
\label{appx:ansvocab}
For our answer vocab, we take any answer that appears in the training set at least $10$ times in the answer annotations. For OKVQA v1.0, our vocabulary size is $2253$ and on v1.1 it is $2250$.

\subsection{Graph Network to \MMBERTBase Baseline}
\label{appx:gntovbbaseline}
Here we more fully describe one of our baselines where we feed the graph network into \MMBERTBase without making a separate prediction.

First, the graph network forward prediction to $G$ is the same as in Sec. 3.2 of the main paper except without the $z^{implicit}$ input as this would make a circular connection between the graph network and \MMBERTAbr. So we take the input symbols and word2vec and we use the graph convolution layers $H^{(l+1)} = f(H^{(l)}, KG)$ where $KG$ is the knowledge graph. As before we end up with $H^{(L)} = G$ which is a $\mathbb{R}^{n \times d_h}$ matrix which corresponds to having a hidden state of size $f_h$ for each node (and therefore concept) in our graph.

Next, we summarize all of these separate hidden states $z_{i}^{symbolic}$ for each node $i$ in the graph. We do this by adding a dummy node and dummy edge type to the input graph where each node in the graph is connected to the dummy node by this dummy edge type. The idea is that we create a special edge type that will try to ``summarize'' the information from all graph hidden states and pass it to this dummy node. We then perform one final RGCN conv layer: $H^{(Summary)} = f(G, KG)$, and extract the hidden state for the dummy node $z_{dummy}^{symbolic}$ or $z_{summary}^{symbolic}$. 

With this summary embedding $z_{summary}^{symbolic}$, we then add this summary vector as an additional input to the \MMBERTAbr model. We compute a linear embedding layer for this input to processes the graph summary vector and make it the same input size as the other transformer inputs. We then append this to the inputs of the \MMBERTAbr. 

We tried other methods to get a single vector representation for the graph network, including a self-attention mechanism, and the self-attention mechanism for only these subset of hidden states (only question and image nodes, only answer nodes etc.). All of these performed worse than this particular way of summarizing the graph network output into one vector.

\section{Network / Training Hyperparameters}
\label{appx:hyper}
Here we record the network and training parameters.
In Table~\ref{table:mmbertparams} we show the network parameters for the \MMBERTAbr baseline and subpart.
In Table~\ref{table:graphparams} we show the network parameters for the Graph Network. 
And in Table~\ref{table:trainparams} we show the training meta-parameters used to train all models.

\begin{table}[h]
\begin{center}
\begin{tabular}{@{}lc@{}}
\toprule
Parameter & Value\\ \midrule
Hidden Size & $768$ \\
Visual Embedding Dim & $2048$ \\
Num Hidden & $12$ \\
Num Attention Heads & $12$ \\
Hidden Dropout Prob & $0.1$ \\
Transfer function & ReLU \\
BERT model name & bert-base-uncased \\
\bottomrule
\end{tabular}
\end{center}
\caption{\MMBERTBase Hyperparameters}
\label{table:mmbertparams}
\end{table}

\begin{table}[h]
\begin{center}
\begin{tabular}{@{}lc@{}}
\toprule
Parameter & Value\\ \midrule
Node Hidden Size & $128$ \\
Num Conv Layers & $2$ \\
Graph Conv Type & RGCN \\
Transfer function & ReLU \\
\MMBERTBase input compress dim & $128$ \\
\bottomrule
\end{tabular}
\end{center}
\caption{Graph Network Hyperparameters}
\label{table:graphparams}
\end{table}

\begin{table}[h]
\begin{center}
\begin{tabular}{@{}lc@{}}
\toprule
Parameter & Value\\ \midrule
Optimizer & AdamW~\cite{kingma2014adam} \\
Scheduler & Warmup Cosine \\
Batch Size & $56$ \\
Learning Rate & $5e-5$ \\
Eps & $1e-8$ \\
Weight Decay & $0$ \\
Warmup Steps & $2000$ \\
Training Steps & $88000$ \\
\bottomrule
\end{tabular}
\end{center}
\caption{Training Hyperparameters}
\label{table:trainparams}
\end{table}

\section{Variance Values for Tables}
\label{appx:variance}
Here we show the sample standard deviations for the runs in our tables in Table~\ref{table:AllAblationStd} and Table~\ref{table:GraphAnalysisstd}.

\begin{table*}[h]
\begin{center}
\begin{tabular}{@{}rlcc@{}}
\toprule
&Method & accuracy & std\\ \midrule
1. & \ModelName (ours) & \bf{32.31}  & 0.24 \\
\hline
\multicolumn{4}{c}{\textbf{Ablation of Symbolic Knowledge}}\\
2. & \MMBERTAbr & 29.26  & 0.76 \\
3. & \ModelName w/ random graph & 30.15  & 0.17  \\
\hline
\multicolumn{4}{c}{\textbf{Ablation of Implicit Knowledge}}\\
4. & \ModelName  w/o BERT pretrain & 26.28  & 0.20  \\
5. & \MMBERTAbr w/o BERT pretrain & 21.82  & 0.34  \\
\hline
\multicolumn{4}{c}{\textbf{Ablation of Network Architecture}}\\
6. & \ModelName no late fusion & 31.10 & 0.12  \\
7. & \ModelName no \MMBERTAbr input & 31.10 & 1.41  \\
8. & \ModelName no \MMBERTAbr input or late fusion & 25.00 & 1.83 \\
9. & \ModelName no backprop into \MMBERTAbr & 27.98 & 1.23 \\
10. & \ModelName with GCN & 30.58 & 0.52  \\
11. & \ModelName feed graph into \MMBERTAbr & 30.99 & 0.16  \\
\hline
\multicolumn{4}{c}{\textbf{Ablation of Graph Inputs}}\\
12. & \ModelName no Q to graph & 31.74 & 0.31 \\
13. & \ModelName no I to graph & 31.59 & 0.34 \\
14. & \ModelName no symbol input & 30.26 & 1.30 \\
15. & \ModelName no w2v & 31.95 & 0.12 \\
\bottomrule
\end{tabular}
\end{center}
\caption{\ModelName ablation on OK-VQA v1.1, with sample standard deviations. Mirrors Table~\ref{table:AllAblation} in the main text.}
\label{table:AllAblationStd}
\end{table*}

\begin{table*}[h]
\begin{center}
\begin{tabular}{@{}rlcc@{}}
\toprule
&Method &  accuracy &  std\\ 
\hline
1. & \ModelName $\max(y^{implicit},y^{symbolic})$ (ours) & \bf{32.31} & 0.24 \\
2. & \ModelName $y^{implicit}$ & 31.47 & 0.05 \\
3. & \ModelName $y^{symbolic}$ & 29.36 & 0.50 \\
4. & \ModelName no backprop $y^{implicit}$ & 28.19 & 1.17 \\
\hline
5. & \ModelName $\text{oracle}(y^{implicit}|y^{symbolic})$  & 36.71 & 0.29\\
\bottomrule
\end{tabular}
\end{center}
\caption{\ModelName Subpart Analysis on OK-VQA v1.1, with sample standard deviations. Mirrors Table~\ref{table:GraphAnalysis} in the main text.}
\label{table:GraphAnalysisstd}
\end{table*}

\section{Additional Ablations}
\label{appx:ablations}
We show the results of two final sets of ablations here.

\begin{table}[h]
\begin{center}
\begin{tabular}{@{}lcc@{}}
\toprule
Method & accuracy & std\\ \midrule
\ModelName (ours) & \bf{32.31} & 0.24 \\
\ModelName DBPedia graph & 31.69 & 1.19\\
\ModelName VG graph & 30.62 & 0.20 \\
\ModelName hasPart KB graph & 30.68 & 0.59\\
\ModelName ConceptNet graph & 31.82 & 0.37\\
\bottomrule
\end{tabular}
\end{center}
\caption{Knowledge Graph Ablation}
\label{table:graphablation}
\end{table}

First in Table~\ref{table:graphablation} we ablate which sources knowledge graphs we use. We show at the top our normal result where we have all $4$ knowledge graph sources. Below that we have the accuracies for just the DBPedia graph, just the VisualGenome graph, just the hasPart KB graph and just the ConceptNet graph. As you might expect, all of these ablations get lower numbers than the combined graph. The two best graphs from this analysis seem to be DBPedia and ConceptNet.

\begin{table}[h]
\begin{center}
\begin{tabular}{@{}lcc@{}}
\toprule
Method & accuracy & std\\ \midrule
\ModelName (ours) & \bf{32.31} & 0.24 \\
\ModelName ImageNet Symbols Only & 31.68 & 0.23 \\
\ModelName Places Symbols Only & 31.47 & 0.27 \\
\ModelName LVIS Symbols Only & 31.48 & 0.39 \\
\ModelName VG Symbols Only & 31.95 & 0.52\\
\bottomrule
\end{tabular}
\end{center}
\caption{Image Symbol Ablation}
\label{table:imgablation}
\end{table}

Next in Table~\ref{table:imgablation} we ablate which image classifiers (and thus which symbols) we use as input to our graph network. At the top we show the full results with all $4$ sets of symbols. Then we individually show the results if we only use the ImageNet symbols, if we only use the Places symbols, the LVIS symbols and the VisualGenome symbols. Again, we see that using any one of these image classifiers rather than all $4$ performs worse than our final method, although the difference between them is not huge small. Based on this experiment, VisualGenome detections were the most significant inputs to the graph network.

\section{More Qualitative Examples}
\label{appx:qual}
\begin{figure*}[t]
\centering
\includegraphics[width=0.88\linewidth]{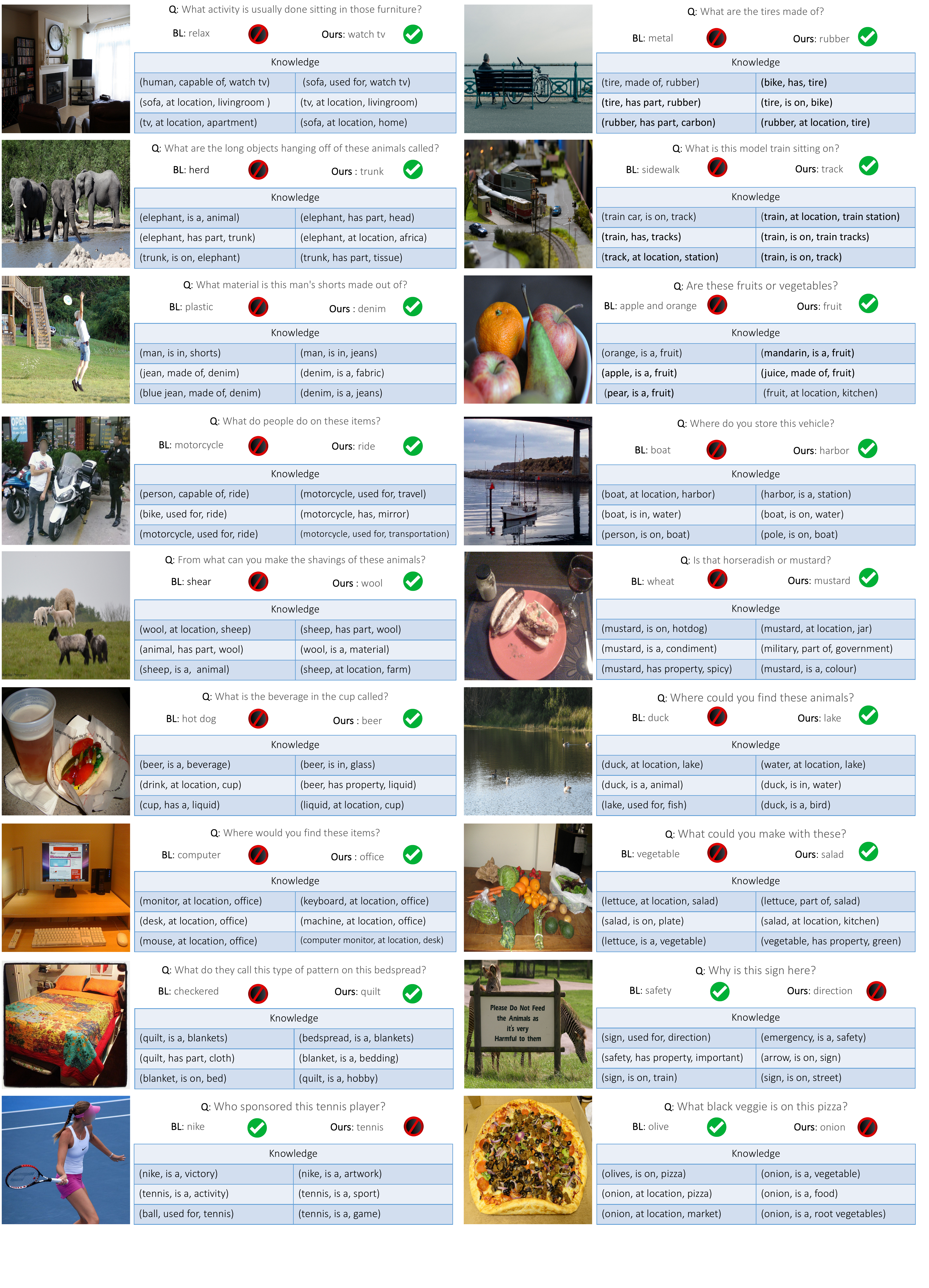}
\caption{More qualitative examples from \ModelName. Showing predictions by our model and the implicit knowledge baseline \MMBERTBase. We show the question, image, and answers given by both models. We also show knowledge in the graph related to the question, answers or image that seemed most relevant.}
\label{fig:qualitativesupp}
\end{figure*}

Finally we show additional qualitative examples in Fig.~\ref{fig:qualitativesupp}. 

\end{document}